\documentclass[review]{elsarticle}

\usepackage{booktabs}
\usepackage{multirow}
\usepackage{makecell}

\usepackage{graphicx}
\usepackage{subcaption}

\usepackage{tabularx}

\usepackage{multirow}

\usepackage{amsfonts}
\usepackage{bbm}

\usepackage{lineno,hyperref}
\modulolinenumbers[5]

\begin{document}

\begin{frontmatter}

\title{Lightweight Modeling of User Context Combining Physical and Virtual Sensor Data}

\author[add1]{Mattia G. Campana\corref{cor1}}
\ead{m.campana@iit.cnr.it}

\author[add1]{Franca Delmastro}
\author[add2]{Dimitris Chatzopoulos}
\author[add2]{Pan Hui}

\cortext[cor1]{Corresponding author}

\address[add1]{Institute for Informatics and Telematics of the National Research Council of Italy (IIT-CNR), Via Giuseppe Moruzzi, 1 56124 Pisa, Italy}

\address[add2]{Hong Kong University of Science and Technology (HKUST), Clear Water Bay, Hong Kong}

\begin{abstract}
The multitude of data generated by sensors available on users' mobile devices, combined with advances in machine learning techniques, support context-aware services in recognizing the current situation of a user (i.e., physical context) and optimizing the system's personalization features. However, context-awareness performances mainly depend on the accuracy of the context inference process, which is strictly tied to the availability of large-scale and labeled datasets.
In this work, we present a framework developed to collect datasets containing heterogeneous sensing data derived from personal mobile devices.
The framework has been used by 3 voluntary users for two weeks, generating a dataset with more than 36K samples and 1331 features. We also propose a lightweight approach to model the user context able to efficiently perform the entire reasoning process on the user mobile device. To this aim, we used six dimensionality reduction techniques in order to optimize the context classification. Experimental results on the generated dataset show that we achieve a 10x speed up and a feature reduction of more than 90\% while keeping the accuracy loss less than 3\%.
\end{abstract}


\end{frontmatter}


\section{Introduction}

The pervasive nature of smart mobile devices, and the plethora of sensors they are equipped with, enabled the development of context-aware systems that can support a variety of personalized services. Example applications include healthcare and well-being~\cite{pierleoni2014android}, human activity recognition~\cite{zhan2014multi},  intelligent transportation systems~\cite{VAHDATNEJAD201643}, personal assistant development~\cite{sun2016contextual}, IoT environments~\cite{perera2014context}, and recommender systems~\cite{Adomavicius2015}. Dey~\textit{et. al.} defined \emph{user context} as the information needed to characterize the situation in which a person is involved~\cite{dey2001understanding}. Consequently, \emph{context-awareness} refers to the ability of a system to adapt and respond proactively to the changes in the users' situation.

\begin{figure}[t]
    \centering
    \includegraphics[width=0.18\columnwidth, angle=90]{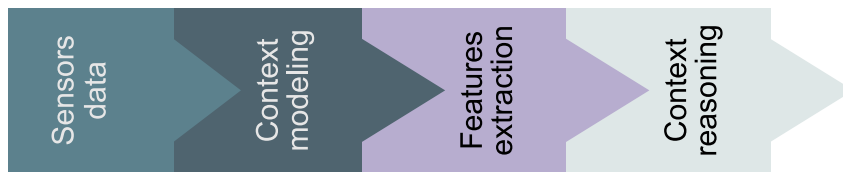}
    \caption{Stages of the user context inference process.}
    \label{fig:context_stages}
\end{figure}

Fig.~\ref{context_stages} depicts the common stages required to infer the user context from raw sensors data~\cite{yurur2016context}.
First, the collected data are modeled and preprocessed to better represent the context information. During this stage, several aspects of the context must be considered (e.g., heterogeneity, comparability, and asynchronous sensors) in order to produce more reliable data for the further processing steps.
Then, a feature vector is built by extracting those information from the sensors data that better characterizes the user context.
Finally, the feature vector is used as input of a context reasoning module that infers more abstract information such as the situation in which the user is currently involved.
The context reasoning module is often represented by a machine learning model (e.g., classifier) trained by using the supervised learning approach.
The performance of this algorithms strongly depends on the availability of a large amount of labeled data. In fact, in order to correctly infer the user context, the learning algorithm requires to process a large set of sensors data (i.e., training set), where each sample is associated with a label that describes its semantic meaning.\\
However, collecting real data for the training phase can be costly or even impractical in some cases, since it requires to develop specific applications to obtain sensors data and then, to manually label them.
In order to cope with this problem, we present \emph{Context Kit} (\emph{CK}), a sensing framework designed to perform large-scale sensing experiments and to easily collect data from real mobile devices.
CK supports the collection of data related to physical sensors (e.g., accelerometer, gyroscope) and to the user interaction with the mobile device (e.g., app usage statistics, phone calls, and connected devices). This information mainly characterizes the user physical context.
Most of the publicly available datasets take into account only a small set of physical sensors. Even though they could be successfully used to recognize standard human activities (e.g., ``running" or ``sitting"), they are not suitable to recognize the user's situation (e.g., ``having a coffee" or ``studying") that is generally characterized by more complex and abstract information.\\
CK allows us to collect heterogeneous context data from real mobile devices. It has been recently used by 3 users for 2 weeks and the collected dataset is composed by 36354 labeled samples, each of them composed by 1331 features.
We publicly release both CK and the collected dataset on the following website: \url{https://contextkit.github.io}.\\
Another challenging problem related to context-aware systems is represented by both context modeling and reasoning stages.
The context reasoning is executed whenever the context of the user needs to be recognized.
Although the model's training can be executed remotely, on powerful servers, the classification phase needs to be executed locally on the mobile device, and for that reason, it has to be lightweight.
Computation offloading mechanisms can be used to perform the classification on remote servers, but this may generate monetary costs (for the use of the server and the data transmission) or poor quality of experience caused by mobile ads.
In addition, the use of remote servers or cloud-based services may demotivate privacy-aware users to use context-aware services, since a third party entity will be in charge of storing users' personal data~\cite{chen2012data,ryan2011cloud}.
Moreover, depending on the communication technology, the delay of the data transmission may make the classification useless with respect to the service optimization, since the user could have changed her context in the meanwhile.
Based on this observations, we argue that the classification process should be executed locally on the mobile devices and produce results in the order of milliseconds.
To this aim, in this paper we also propose a lightweight approach to model the user context in order to speed up the context reasoning process and perform the entire computation on the local device.
Traditional context-aware systems heavily rely on hand-crafted features, which are limited to human domain knowledge and can affect their generalization performance~\cite{bengio2013deep}.
On the contrary, we propose the use of latent features starting from a high number of heterogeneous sensed data to speed up the classification process and allow its local execution on the mobile device.
Latent features represent hidden context patterns modeled as numeric vectors that are extracted from raw sensor data and can be automatically inferred using Dimensionality Reduction (DR) techniques in a data-driven way.
We investigate if it is possible to reach the same accuracy level obtained with the raw features using those extracted by six well-known DR techniques.\\
In summary, our contribution in this paper is threefold.
First, we propose a new framework to perform large-scale sensing experiments and to easily collect raw context data.
Second, we release a real and labeled dataset which contains data from both physical and virtual sensors collected by a set of heterogeneous smartphones.
Third, we propose a new methodology to model the raw context data using latent features in order to speed up the context reasoning process and perform the entire computation on the local device.

\section{Related Work and Main Contribution}
\label{sec:related}
Nowadays, context-awareness is a fundamental requirement to develop ubiquitous and pervasive systems.
The main goal of these systems is to infer the user context from different sensors to provide autonomous and proactive services to the final user.
According to Bettini et. al~\cite{bettini2010survey}, the term "sensor" can describe not only the physical device (e.g., accelerometer or GPS), but also any kind of data source that can be useful to characterize the user context; in the latter case, we refer to this kind of data source as \emph{virtual sensors}~\cite{yurur2016context}.
We can consider as virtual sensors data the application usage statistics, the ringtone level (e.g., if the smartphone is in silent mode, probably the user is busy), and the display status.
In this section we discuss the characteristics of state-of-the-art solutions related to different aspects of the context inference process, highlighting how the combination of physical and virtual sensors improves the process and the advantages introduced by our proposal.

\subsection{Context Information}
Most of the context-aware solutions proposed in the literature focus on few sensor data for inferring the user context. For instance, keeping into account only the accelerometer data is a common practice for the user's gait recognition and fall detection~\cite{abdullah2012towards,hoang2013adaptive,app7101101}, while the GPS coordinates and the list of Wi-Fi access points are commonly used to infer the current location of the user and her social interactions~\cite{chon2012automatically,de2013interdependence}.
Even though a small set of sensors could be sufficient to recognize standard human activities (often requiring complex algorithms), inferring more abstract information like the user's situation requires the combination of several heterogeneous sensors.
To this aim, we propose to characterize the user context using a large set of sensors commonly available on commercial mobile devices. Specifically, we define the user's situation based on the combination of the following physical and virtual sensors: the user's current gait (e.g., walking or on bicycle); time's information (e.g., morning/night, or weekday/weekend); the list of running applications, information related to the device's audio status (e.g., ringtone level, or connected earphones); weather conditions; battery level; if the device is plugged into a power source; list of connected Bluetooth (BT) devices; display status (i.e., on/off, and rotation angle); GPS location; if the device is connected to a Wi-Fi Access Point; physical sensors data (e.g., gyroscope, accelerometer, and light); devices in proximity (both via BT and Wi-Fi Direct); and if the user is using the camera for taking a picture or recording a video.

\subsection{Context Acquisition}
In order to assist the collection of sensing data from commercial mobile devices, several sensing frameworks have been recently proposed in the literature. 
Funf~\cite{Aharony:2011:SFI:2072697.2073099} is one of the first; it allows to easily activate and deactivate sensor monitoring through a configuration file, but its maintenance is discontinued and its support for recent mobile operating systems is very limited.
SensingKit~\cite{Katevas:2014:PSM:2639108.2642910} is a more recent project proposed by the Queen Mary University of London. It is a multi-platform framework (i.e., iOS and Android), but it does not support useful sensors (e.g., Wi-Fi Direct) or to query external services to obtain additional information that can be useful to characterize the user context (e.g., weather conditions).
In order to cope with these limitations, we propose a new sensing framework specially designed to perform large-scale sensing experiments with mobile devices.
Our framework allows data collection from several sensors (both physical and virtual) to characterize the user context and it implements some unique features that are not available in other similar frameworks.
For instance, it supports the usage of different external data sources, such as OpenWeatherMap (\url{https://openweathermap.org}) to download the weather conditions of the geographical area in which the user is located.
In addition, CK exploits both BT 4.0 and Wi-Fi Direct to estimate the proximity between different devices (e.g., smartphones, IoT devices, external sensors or Wi-Fi Access Points). In fact, proximity information can be used to further characterize the user physical context, both in terms of location and possible interactions with other users or devices~\cite{boonstra2015mapping}.

\subsection{Context Modeling}
Raw sensor data must be modeled to infer a more abstract representation of the user context (i.e., the user's situation).
The most challenging part of the context modeling process is the identification of the sensor information that is the most descriptive of the user context~\cite{pejovic2015anticipatory}.
Traditional approaches focus on modeling the context information using software engineering formalisms such as ontology-based technologies (e.g., RDF~\cite{forstadius2005rdf}, graphical models (e.g., UML~\cite{henricksen2003generating} or ORM~\cite{halpin2010object}),  and OWL~\cite{wang2004ontology}), or simple mark-up schemes (e.g., XML~\cite{knappmeyer2010contextml}).
Each of them has several pros and cons in terms of different aspects (e.g., expressive power of the model, complexity), and they can also affect the computation performance and scalability of the context reasoning process~\cite{bettini2010survey, yurur2016context}.
Moreover, defining these models is a time-consuming task and requires to manually identify the best set of features to characterize the user context.
On the other hand, raw data coming directly from sensors may contain irrelevant or redundant features that may affect the performance of the context classification.\\
In order to cope with these drawbacks, instead of using time-consuming models or manually feature selections, we propose to learn a compact representation of raw sensor information in a data-driven way.
More specifically, we use the following DR techniques: (i) Autoencoder (AE)~\cite{hinton2006reducing}, a special class of Artificial Neural Networks that uses hidden layers of neurons to learn a non-linear and compressed representation of the input data; (ii) Non-Negative Matrix Factorization (NMF)~\cite{huang2014non}, which decomposes the original feature space into two non-negative matrices that can be applied to reduce the dimensionality of the original feature space; (iii) Feature Agglomeration (FA)~\cite{5364970}, which uses a recursively approach to merge the features that look similar in a hierarchical way; (iv) Principal Component Analysis (PCA)~\cite{Olive2017}, which exploits the eigenvalue decomposition of the data covariance matrix to suggest an eigenvector along which the projection of data should have the minimal distortion; and (v - vi) Random Projection (either Sparse -SRP- or Gaussian -GRP-)~\cite{Bingham:2001:RPD:502512.502546}, which represent a quicker alternative to the PCA, where the original feature space is projected to a lower subspace using a random kernel matrix.
In this work, we explore the use of these DR techniques to automatically extract the most relevant features from the raw data and to speed up the recognition of the user context in order to perform the entire computation on the local mobile device.

\begin{figure}[t]
    \centering
    \includegraphics[width=0.18\columnwidth]{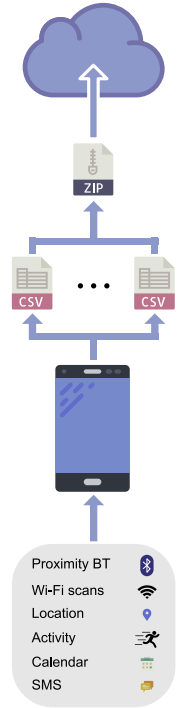}
    \caption{High-level architecture of the Context Kit framework.}
    \label{fig:arch}
\end{figure}

\subsection{Context Reasoning}
Context Reasoning (CR) represents the core of the inference process. 
Most of the CR approaches proposed in the literature use supervised learning algorithms to learn a classifier from raw sensor data like the following: i) k-Nearest Neighbor (k-NN)~\cite{1053964}, which predicts the label of any new sample $x$ by running majority vote among its $k>1$ nearest neighbors; ii) Support Vector Machines (SVM)~\cite{708428}, a binary classifier that learns the hyperplane maximizing the margin between samples of different classes; and iii) Classification and Regression Tree (CART)~\cite{breiman2017classification}, which breaks down the feature space in a way to minimize the misclassification error.
Here, we benchmark these three classifiers both in terms of classification accuracy and execution time.
Moreover, we compare their ability to correctly classify the user context using both the raw sensor data and the features learned by the aforementioned DR techniques.

\subsection{Context Datasets}
The performance of supervised learning algorithms mostly depends on the availability of labeled data~\cite{yurur2016context}.
However, most of the publicly available datasets (e.g., UMA Fall~\cite{casilari2016analysis}, UniMiB SHAR~\cite{app7101101}, and RealWorld (HAR)~\cite{sztyler2016body}) take into account only a limited number of sensors since their main purpose is to assist the creation of models for specific activity recognition (e.g., user's gait and fall detection).
Therefore, these datasets are not suitable for learning a model able to infer the user's current situation.
In this work, we present a new dataset of labeled context data acquired with an Android application created on top of CK sensing framework.
In order to obtain a real dataset, we avoid to perform the data acquisition in controlled environments (e.g., lab), and we enrolled a set of voluntary users equipped with heterogeneous commercial mobile devices, with different characteristics and sensors.
The users signed an informed consent including all the policies adopted for personal data storage, management and analysis, including the publication of the anonymized dataset, according to the EU GDPR regulation.  
\section{Context Kit and Dataset}




CK (depicted in Fig.~\ref{fig:arch}) is designed to collect a high variety of sensors data, configurable depending on the experiment's requirements. To this aim, it implements every sensing category as an independent module, which can be activated or deactivated by using a configuration file.
When CK is running, sensor data are stored in different log files (one for each sensing module) and saved on the device's hard drive.
Finally, in order to simplify the data collection from several devices, CK includes also a network module which compresses and sends the log files to a specified remote endpoint. 



\subsection{Context Labeler and Data acquisition}

Research studies in the area of activity recognition and human behavior modeling base their results on experiments performed in controlled environments (e.g., a research laboratory)~\cite{app7101101}.
During the data collection process (often performed with the same device), some volunteers are asked to perform some activities that have been previously defined by researchers.
However, in the real world we have heterogeneous devices and different users may have different ways of doing the same activity; thus the experimental results usually diverge from those obtained in the lab~\cite{Mafrur2015}.

\begin{figure}[t]
    \centering
    \includegraphics[width=0.2\columnwidth]{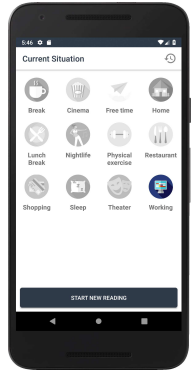}
    \caption{UI of the Context Labeler app.}
    \label{fig:ui}
\end{figure}

In order to build a real dataset, we have developed an Android application called \emph{Context Labeler} that includes the CK framework as a library and allows to associate predefined labels with the collected context data.
We asked three volunteers to install Context Labeler on their own smartphones and to select their daily life activities from the list of predefined labels (depicted in Fig. 3).
After the activity selection, Context Labeler starts the CK framework that monitors the context data unobtrusively in the background.
When the current activity ends, the user manually stops the data reading using a specific button, and both the sensed data and the selected label are stored into the device's hard drive.\\
By using Context Labeler, we collected the context data from the following devices: a Nexus 5 with Android 6.0.1, a Xiaomi Mi 5 with Android 7.1.2, and a Reader P10 with Android 6.0.
In addition, to avoid introducing biases during the data acquisition, we did not define any constraints for the user behavior during the experiment.
On the contrary, we encouraged the volunteers to use their smartphones without worrying about the positions of the device (e.g., trousers' pockets, or hand).

\begin{figure}[t]
	\centering
    \includegraphics[width=0.6\textwidth]{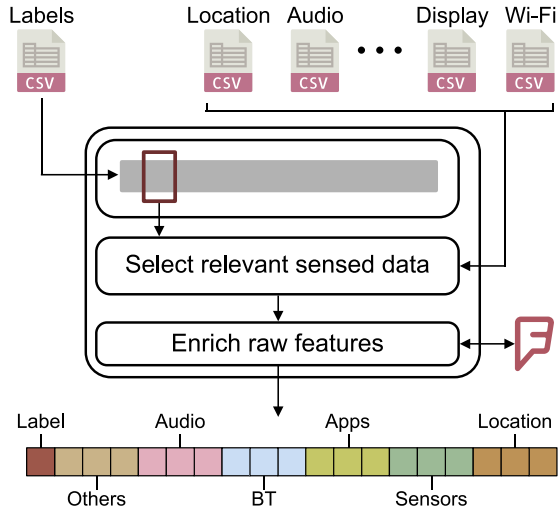}
    \caption{Scheme of the data processing used to generate the final dataset.}
    \label{fig:data_processing}
\end{figure}

\subsection{Dataset Generation}

Different sensors or events monitored by CK may have different sampling rates. Therefore, even if two different sensor data refer to the same situation, they may have different timestamps.
Moreover, each label collected by Context Labeler is stored, together with its duration, in a separated log file.
In order to generate the data samples required for the context reasoning phase, we process the log files as depicted in Fig.~\ref{fig:data_processing}.
First, we use a sliding window to split the duration of each activity into slots of 1 second each.
Second, for every time slot, we fetch from the log files only the sensor data with the closest reading timestamp to the starting time of the current slot.
Then, we enrich the raw features with additional categorical information.
For example, using the Foursquare APIs (\url{https://developer.foursquare.com}), we extend the location features retrieving the category of the most probable venue according to the GPS coordinates.
Finally, we represent the categorical features (e.g., location category) using the well-known \emph{One Hot Encoding} technique, and we combine them with the continuous features derived from physical sensor values, by associating the corresponding situation's label.
The resulting dataset contains 36354 labeled samples, with 1331 features per sample.

\section{Evaluation}
\label{sec:eval}

The main goal of the experimental evaluation is to find a combination of classifier and DR technique that can rapidly detect the user physical context with a high accuracy, while performing the entire computation on the local device. To this aim, we run four different experiments.
In the first experiment, we measure the classification accuracy of the three classifiers.
We initially calculate the accuracy by using the raw feature vectors.
Then, we examine the accuracy of the classifiers after the DR techniques application to automatically select the most relevant features and reduce the size of the data samples.
In the second experiment, we measure the time each DR technique needs to reduce the dimensions of all the data samples in the dataset. Then, we calculate the time required to train and test each classifier by using the DR techniques and the raw data.
Finally, we perform a subject-independent cross-validation, in which the training set consists of the dataset samples generated by users not included in the test set~\cite{app7101101}.

\begin{figure}[t]
    \centering
    \begin{subfigure}{.49\textwidth}
        \centering
        \includegraphics[width=\linewidth, angle=270]{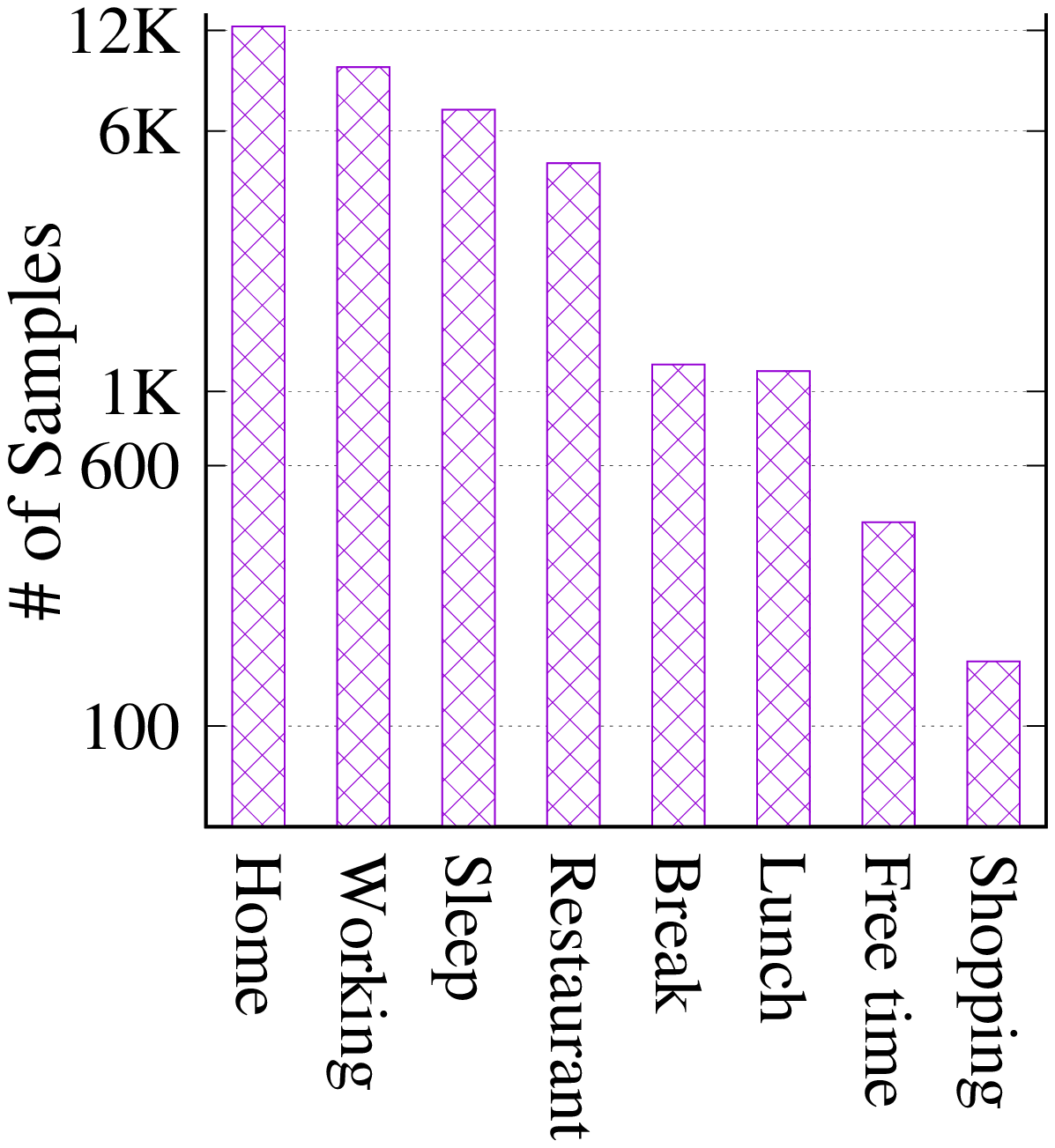}
        \caption{}
        \label{fig:loc_all}
    \end{subfigure}
    \begin{subfigure}{.49\textwidth}
        \centering
        \includegraphics[width=\linewidth, angle=270]{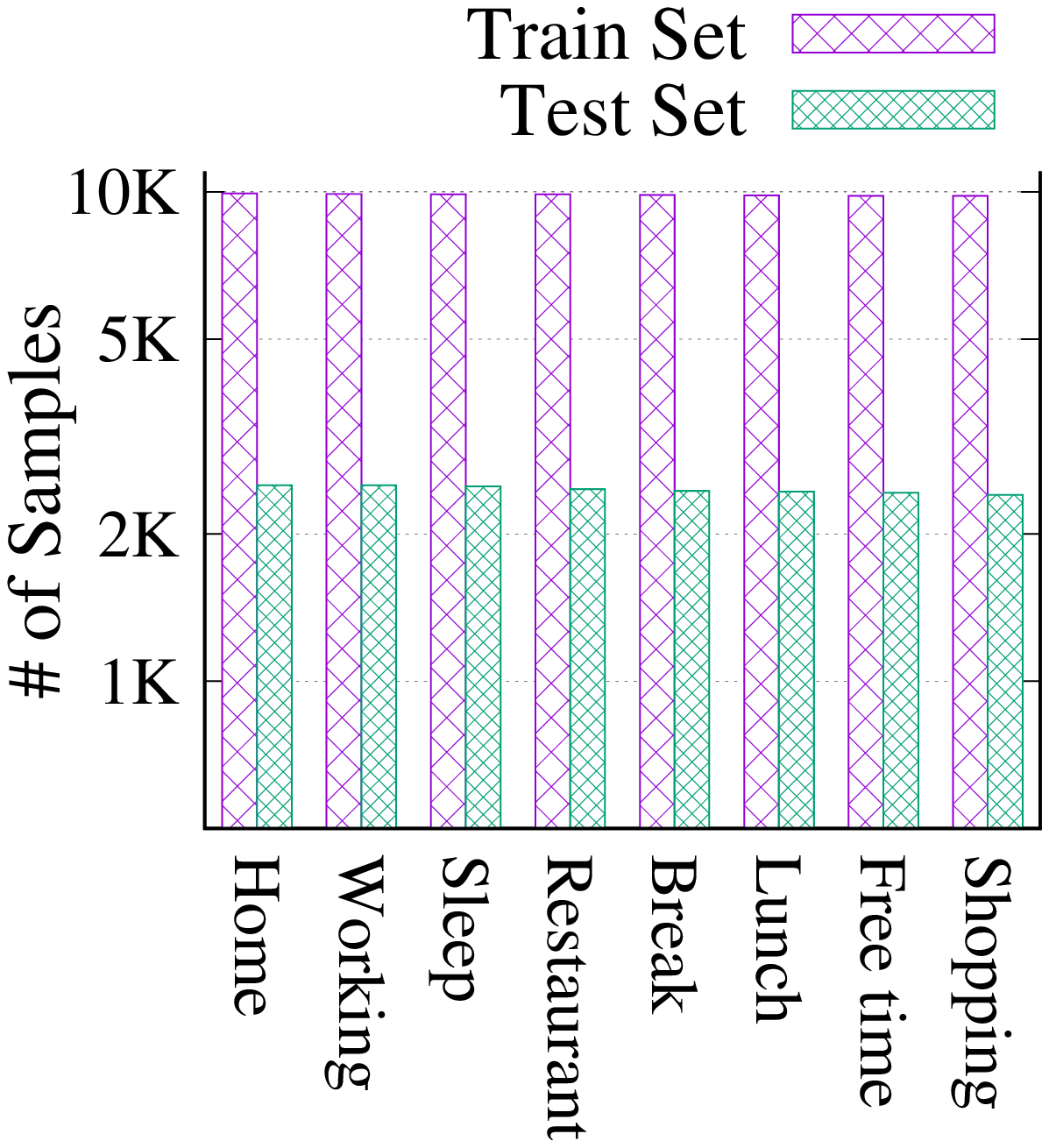}
        \caption{}
        \label{fig:loc_exp_area}
    \end{subfigure}

    \caption{Distribution of the labels in our dataset (a), and Distribution of the train and the test labels after the oversampling (b).}
    \label{fig:dist}
\end{figure}

\subsection{Dataset balance}

Since we are proposing a supervised approach, we first process the collected dataset to produce the train and test sets.
Fig. 5 shows the labels distribution of the data samples in our dataset.
It is worth noting that the distribution is highly skewed, where some classes appear more frequently than others.
This problem is a well-studied phenomenon in the literature, known as the ``Imbalanced dataset problem'' and it appears in many real-world datasets~\cite{Krawczyk2016}.
Imbalanced datasets can negatively influence the generalization and reliability of supervised learning algorithms, since the resulting classifiers may be biased by the majority classes.
In order to balance our dataset, we employ the SMOTE algorithm~\cite{chawla2002smote}, an oversampling technique that creates new synthetic data samples in the minority classes, varying the features values of the existing data points based on their $k$ nearest neighbors in the feature space.
After the dataset balancing, we randomly split the dataset in training (80\%, 78841 samples) and test (20\%, 19711 samples) sets.
Fig. 6 shows the distributions of the train and test sets after the oversampling.

\subsection{Classification Accuracy}

In the first experiment we compare the six DR techniques, in terms of classification accuracy, when used as preprocessing step on the three selected classifiers.
As a first step, we calculate the accuracy level obtained by raw features and we note that all classifiers reach about 99\% accuracy by using the raw data.\\
Then, we test the combination of our models with the six DR approaches in order to evaluate the impact of using latent features on the overall accuracy.\\
Fig. 7 shows the classification accuracy of each classifier for a different number of latent features produced by the six DR techniques.
It is worth noting that k-NN and CART achieve similar results to those obtained with raw data by employing a limited number of latent features produced by DR techniques, while SVM performs at least 10\% worse.

\subsection{DR Time Performance}

Continuing the previous experiment we examine the time needed by six DR techniques to produce the latent features by using all the samples in our dataset.
According to the results depicted in Fig. 8, we can classify the DR techniques in three categories: i) SRP and GRP run in milliseconds, ii) AE and PCA in a few seconds, and iii) FA and NMF in minutes.
It is worth noting that the time needed by NMF increases with the number of latent features, while the others have nearly a constant execution time.

\subsection{Classification Time Performance}

Based on the results obtained in the first experiment, we select for each DR technique the number of features which allows each classifier to obtain the best classification accuracy.
For instance, the number of latent features for CART and AE is 25, with 97\% of classification accuracy. Then, by using these values, we compared the three classifiers in terms of execution times for both the train and test phases.
We present the produced results in Fig.~\ref{fig:classifier_exec_time}. As expected, the classifiers have their worst performance on the raw data (with k-NN requiring more than 16 minutes for testing) since they are considering more features.
k-NN maintains the higher execution times (less than 1 minute) for the test phase, while performing better in the training phase. 
Instead, CART achieves the best performance in testing even though it requires more time for training, but still in the order of few seconds. 

\begin{figure}[t]
    \centering
    \begin{subfigure}{.32\linewidth}
        \centering
        \includegraphics[width=\linewidth, angle=270]{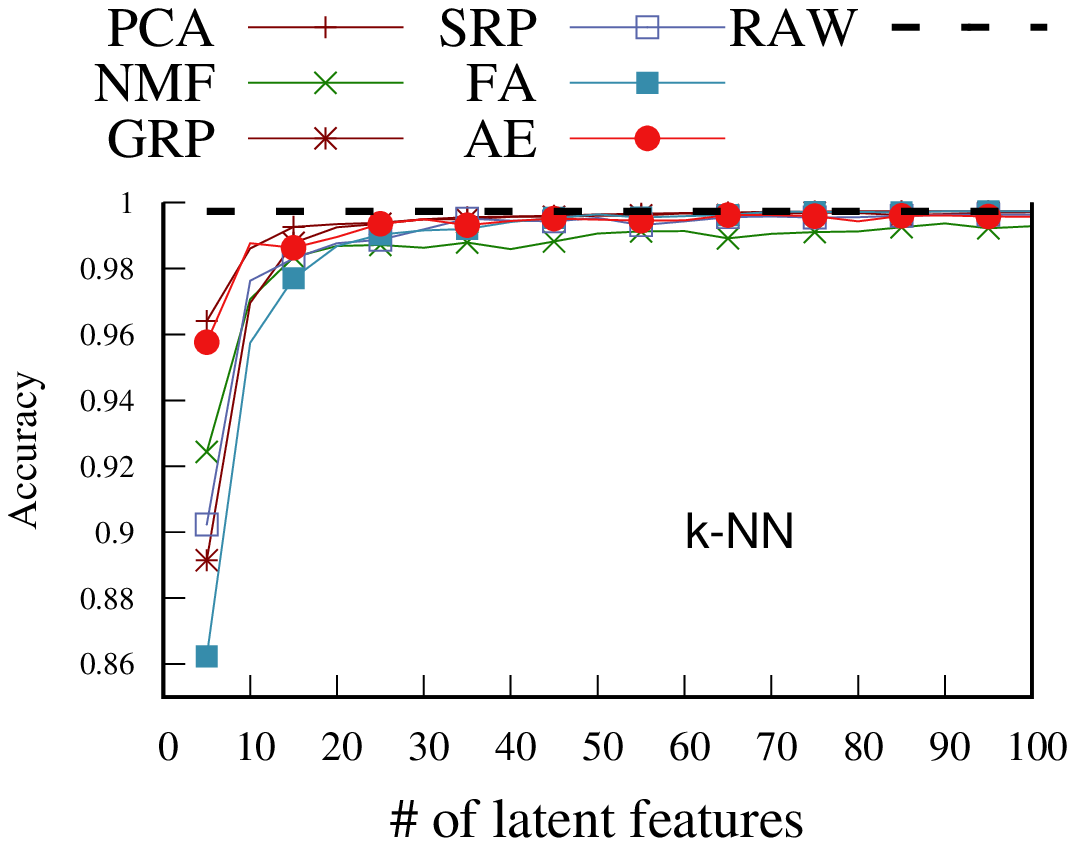}
        \caption{}
        \label{}
    \end{subfigure}
    \begin{subfigure}{.32\linewidth}
        \centering
        \includegraphics[width=\linewidth, angle=270]{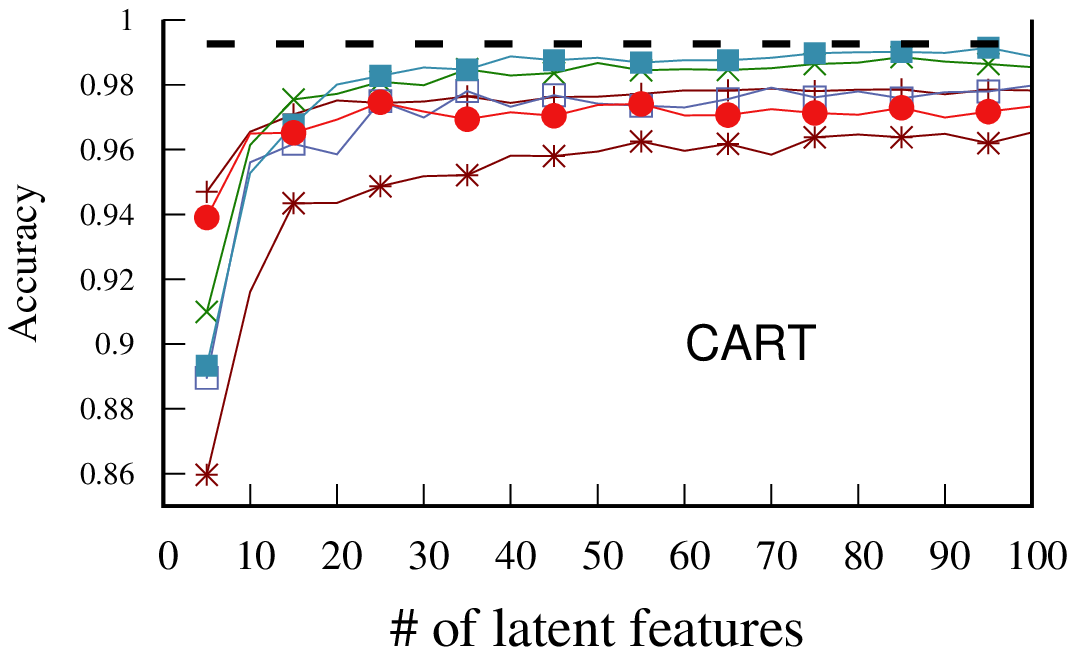}
        \caption{}
        \label{}
    \end{subfigure}\\
    \begin{subfigure}{.32\linewidth}
        \centering
        \includegraphics[width=\linewidth, angle=270]{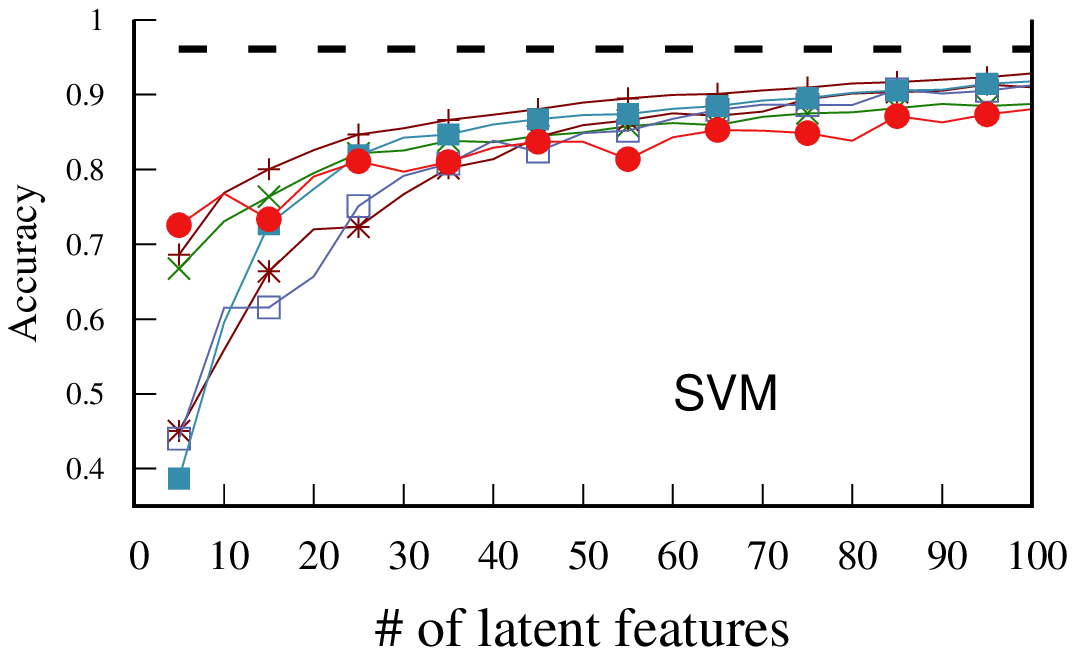}
        \caption{}
        \label{}
    \end{subfigure}

    \caption{Classification accuracy with raw and latent features.}
    \label{fig:classification_acc}
\end{figure}

\begin{figure}[t]
    \centering
    \includegraphics[width=.6\columnwidth,angle=270]{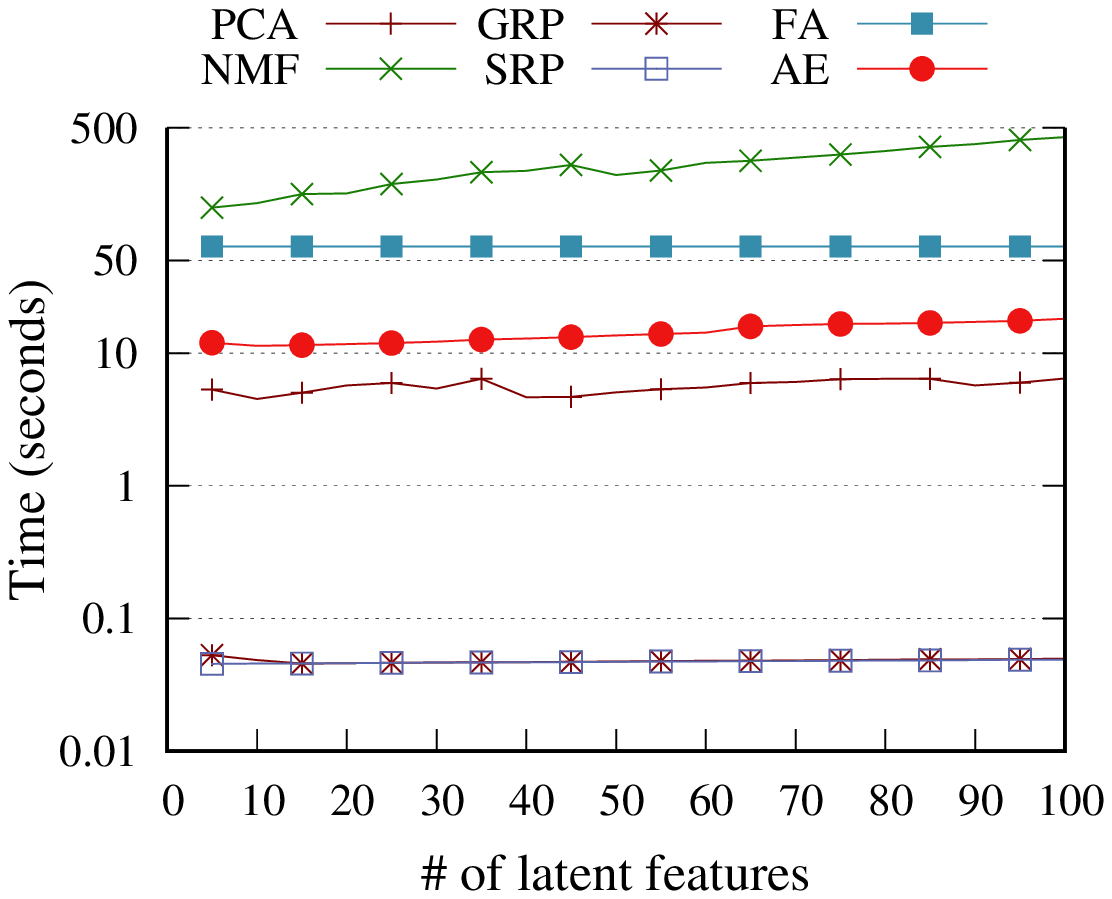}
    \caption{Dimensionality reduction execution time.}
    \label{fig:dim_red_time}
\end{figure}

\textbf{Subject-Independent Cross-Validation Accuracy} \\
The main goal of this experiment is to evaluate the expressive power of the latent features. In other words, we want to test if the latent features extracted by the DR techniques are able to generalize the different contexts, independently of the user who generated the training data.
To this aim, the training set is made of dataset samples generated by users not included in the test set~\cite{app7101101}.
From the results depicted in Fig.~\ref{fig:si_class_accuracy} is evident the accuracy reduction with respect to the first experiment due to the characteristics of the supervised learning approach.
However, even though all classifiers have lost about 50\% of accuracy, they still perform approximately 3 times better than a Random Guesser (i.e., a classifier who randomly predicts the label).
Interestingly, in this experiment all the classifiers perform better using the latent features instead of the raw data.
This proves the capability of the DR techniques to infer a set of features able to correctly represent single situations independently of the behavioral patterns of the specific users.






\section{Conclusions and future work}
\label{sec:conclusions}
In this work, we present a new framework to collect physical and virtual sensor data from personal mobile devices. We have used the framework to collect a real dataset composed of more 36K samples and 1331 features.
Based on this dataset, we have conducted four experiments to evaluate a lightweight approach to model the user context and speed up the context reasoning process.
Experimental results show that the entire context reasoning process can be performed on the local mobile device by appropriately selecting a classifier and dimensionality reduction techniques. We are planning to integrate the proposed approach in a middleware solution we defined for the development of context-aware recommender systems in mobile opportunistic environments~\cite{ARNABOLDI20173,7509388,8024034,Chatzopoulos:2016:RRR:2964284.2967233}. Here, mobile systems and applications can rely only on the local information provided by user mobile devices and intermittent wireless connectivity among them. Therefore, users' context is highly dynamic and wireless communications among mobile devices to exchange and share data are limited in time due to users' mobility. In this scenario, a lightweight modeling of user physical context will improve the personalization of recommendations.  

\begin{figure}[t]
	\centering
    \includegraphics[height=\columnwidth,angle=270]{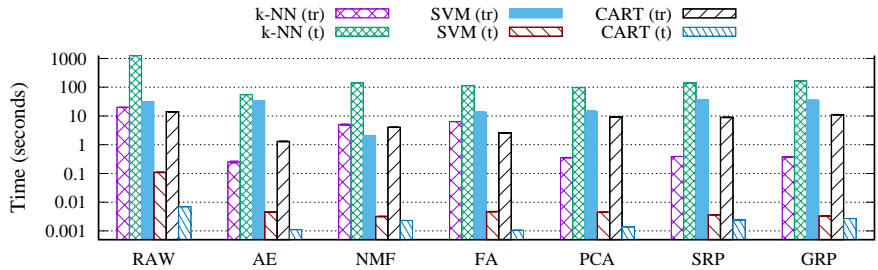}
    \vspace{-0.5cm}
    \caption{Classifiers execution times.\vspace{-0.3cm}}
    \label{fig:classifier_exec_time}
\end{figure}

\begin{figure}[t]
	\centering
    \includegraphics[height=\columnwidth,angle=270]{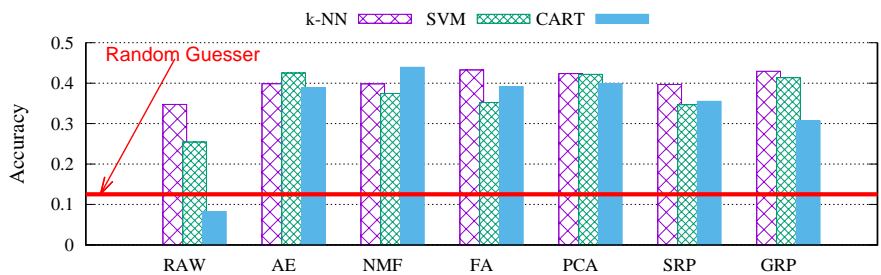}
    \vspace{-0.5cm}
    \caption{Subject-Independent Classification accuracy.\vspace{-0.3cm}}
    \label{fig:si_class_accuracy}
\end{figure}

\section{Acknowledgements}
The authors would like to thank Reza Hadi Mogavi for his useful suggestions related to the machine learning techniques used in this work. This research was carried out in the framework of the INTESA project (CUP CIPE D78I16000010008), co-funded by the Tuscany Region (Italy) and MIUR under the Programme FAR FAS 2007-2013, and by projects 26211515 and 16214817 from the Research Grants Council of Hong Kong.

\bibliography{paper}

\begin{thebibliography}{10}
\expandafter\ifx\csname url\endcsname\relax
  \def\url#1{\texttt{#1}}\fi
\expandafter\ifx\csname urlprefix\endcsname\relax\def\urlprefix{URL }\fi
\expandafter\ifx\csname href\endcsname\relax
  \def\href#1#2{#2} \def\path#1{#1}\fi

\bibitem{pierleoni2014android}
P.~Pierleoni, L.~Pernini, A.~Belli, L.~Palma, An android-based heart monitoring
  system for the elderly and for patients with heart disease, International
  journal of telemedicine and applications 2014 (2014) 10.

\bibitem{zhan2014multi}
K.~Zhan, S.~Faux, F.~Ramos, Multi-scale conditional random fields for
  first-person activity recognition, in: Pervasive Computing and Communications
  (PerCom), 2014 IEEE International Conference on, IEEE, 2014, pp. 51--59.

\bibitem{VAHDATNEJAD201643}
H.~Vahdat-Nejad, A.~Ramazani, T.~Mohammadi, W.~Mansoor,
  \href{http://www.sciencedirect.com/science/article/pii/S2214209616000036}{A
  survey on context-aware vehicular network applications}, Vehicular
  Communications 3 (2016) 43 -- 57.
\newblock \href {https://doi.org/https://doi.org/10.1016/j.vehcom.2016.01.002}
  {\path{doi:https://doi.org/10.1016/j.vehcom.2016.01.002}}.
\newline\urlprefix\url{http://www.sciencedirect.com/science/article/pii/S2214209616000036}

\bibitem{sun2016contextual}
Y.~Sun, N.~J. Yuan, Y.~Wang, X.~Xie, K.~McDonald, R.~Zhang, Contextual intent
  tracking for personal assistants, in: Proceedings of the 22nd ACM SIGKDD
  International Conference on Knowledge Discovery and Data Mining, ACM, 2016,
  pp. 273--282.

\bibitem{perera2014context}
C.~Perera, A.~Zaslavsky, P.~Christen, D.~Georgakopoulos, Context aware
  computing for the internet of things: A survey, IEEE communications surveys
  \& tutorials 16~(1) (2014) 414--454.

\bibitem{Adomavicius2015}
G.~Adomavicius, A.~Tuzhilin,
  \href{https://doi.org/10.1007/978-1-4899-7637-6_6}{Context-Aware Recommender
  Systems}, Springer US, Boston, MA, 2015, pp. 191--226.
\newblock \href {https://doi.org/10.1007/978-1-4899-7637-6_6}
  {\path{doi:10.1007/978-1-4899-7637-6_6}}.
\newline\urlprefix\url{https://doi.org/10.1007/978-1-4899-7637-6_6}

\bibitem{dey2001understanding}
A.~K. Dey, Understanding and using context, Personal and ubiquitous computing
  5~(1) (2001) 4--7.

\bibitem{yurur2016context}
{\"O}.~Y{\"u}r{\"u}r, C.~H. Liu, Z.~Sheng, V.~C. Leung, W.~Moreno, K.~K. Leung,
  Context-awareness for mobile sensing: A survey and future directions, IEEE
  Communications Surveys \& Tutorials 18~(1) (2016) 68--93.

\bibitem{chen2012data}
D.~Chen, H.~Zhao, Data security and privacy protection issues in cloud
  computing, in: Computer Science and Electronics Engineering (ICCSEE), 2012
  International Conference on, Vol.~1, IEEE, 2012, pp. 647--651.

\bibitem{ryan2011cloud}
M.~D. Ryan, Cloud computing privacy concerns on our doorstep, Communications of
  the ACM 54~(1) (2011) 36--38.

\bibitem{bengio2013deep}
Y.~Bengio, Deep learning of representations: Looking forward, in: International
  Conference on Statistical Language and Speech Processing, Springer, 2013, pp.
  1--37.

\bibitem{bettini2010survey}
C.~Bettini, O.~Brdiczka, K.~Henricksen, J.~Indulska, D.~Nicklas,
  A.~Ranganathan, D.~Riboni, A survey of context modelling and reasoning
  techniques, Pervasive and Mobile Computing 6~(2) (2010) 161--180.

\bibitem{abdullah2012towards}
S.~Abdullah, N.~D. Lane, T.~Choudhury, Towards population scale activity
  recognition: A framework for handling data diversity., in: AAAI, 2012.

\bibitem{hoang2013adaptive}
T.~Hoang, T.~D. Nguyen, C.~Luong, S.~Do, D.~Choi, Adaptive cross-device gait
  recognition using a mobile accelerometer., JIPS 9~(2) (2013) 333.

\bibitem{app7101101}
D.~Micucci, M.~Mobilio, P.~Napoletano,
  \href{http://www.mdpi.com/2076-3417/7/10/1101}{Unimib shar: A dataset for
  human activity recognition using acceleration data from smartphones}, Applied
  Sciences 7~(10) (2017).
\newblock \href {https://doi.org/10.3390/app7101101}
  {\path{doi:10.3390/app7101101}}.
\newline\urlprefix\url{http://www.mdpi.com/2076-3417/7/10/1101}

\bibitem{chon2012automatically}
Y.~Chon, N.~D. Lane, F.~Li, H.~Cha, F.~Zhao, Automatically characterizing
  places with opportunistic crowdsensing using smartphones, in: Proceedings of
  the 2012 ACM Conference on Ubiquitous Computing, ACM, 2012, pp. 481--490.

\bibitem{de2013interdependence}
M.~De~Domenico, A.~Lima, M.~Musolesi, Interdependence and predictability of
  human mobility and social interactions, Pervasive and Mobile Computing 9~(6)
  (2013) 798--807.

\bibitem{Aharony:2011:SFI:2072697.2073099}
N.~Aharony, W.~Pan, C.~Ip, I.~Khayal, A.~Pentland,
  \href{http://dx.doi.org/10.1016/j.pmcj.2011.09.004}{Social fmri:
  Investigating and shaping social mechanisms in the real world}, Pervasive
  Mob. Comput. 7~(6) (2011) 643--659.
\newblock \href {https://doi.org/10.1016/j.pmcj.2011.09.004}
  {\path{doi:10.1016/j.pmcj.2011.09.004}}.
\newline\urlprefix\url{http://dx.doi.org/10.1016/j.pmcj.2011.09.004}

\bibitem{Katevas:2014:PSM:2639108.2642910}
K.~Katevas, H.~Haddadi, L.~Tokarchuk,
  \href{http://doi.acm.org/10.1145/2639108.2642910}{Poster: Sensingkit: A
  multi-platform mobile sensing framework for large-scale experiments}, in:
  Proceedings of the 20th Annual International Conference on Mobile Computing
  and Networking, MobiCom '14, ACM, New York, NY, USA, 2014, pp. 375--378.
\newblock \href {https://doi.org/10.1145/2639108.2642910}
  {\path{doi:10.1145/2639108.2642910}}.
\newline\urlprefix\url{http://doi.acm.org/10.1145/2639108.2642910}

\bibitem{boonstra2015mapping}
T.~W. Boonstra, M.~E. Larsen, H.~Christensen, Mapping dynamic social networks
  in real life using participants' own smartphones, Heliyon 1~(3) (2015)
  e00037.

\bibitem{pejovic2015anticipatory}
V.~Pejovic, M.~Musolesi, Anticipatory mobile computing: A survey of the state
  of the art and research challenges, ACM Computing Surveys (CSUR) 47~(3)
  (2015) 47.

\bibitem{forstadius2005rdf}
J.~Forstadius, O.~Lassila, T.~Seppanen, Rdf-based model for context-aware
  reasoning in rich service environment, in: Pervasive Computing and
  Communications Workshops, 2005. PerCom 2005 Workshops. Third IEEE
  International Conference on, IEEE, 2005, pp. 15--19.

\bibitem{henricksen2003generating}
K.~Henricksen, J.~Indulska, A.~Rakotonirainy, Generating context management
  infrastructure from high-level context models, in: In 4th International
  Conference on Mobile Data Management (MDM)-Industrial Track, Citeseer, 2003.

\bibitem{halpin2010object}
T.~Halpin, Object-role modeling: Principles and benefits, International Journal
  of Information System Modeling and Design (IJISMD) 1~(1) (2010) 33--57.

\bibitem{wang2004ontology}
X.~H. Wang, D.~Q. Zhang, T.~Gu, H.~K. Pung, Ontology based context modeling and
  reasoning using owl, in: Pervasive Computing and Communications Workshops,
  2004. Proceedings of the Second IEEE Annual Conference on, Ieee, 2004, pp.
  18--22.

\bibitem{knappmeyer2010contextml}
M.~Knappmeyer, S.~L. Kiani, C.~Fr{\`a}, B.~Moltchanov, N.~Baker, Contextml: A
  light-weight context representation and context management schema, in:
  Wireless Pervasive Computing (ISWPC), 2010 5th IEEE International Symposium
  on, IEEE, 2010, pp. 367--372.

\bibitem{hinton2006reducing}
G.~E. Hinton, R.~R. Salakhutdinov, Reducing the dimensionality of data with
  neural networks, science 313~(5786) (2006) 504--507.

\bibitem{huang2014non}
K.~Huang, N.~D. Sidiropoulos, A.~Swami, Non-negative matrix factorization
  revisited: Uniqueness and algorithm for symmetric decomposition, IEEE
  Transactions on Signal Processing 62~(1) (2014) 211--224.

\bibitem{5364970}
P.~M. Ciarelli, E.~Oliveira, Agglomeration and elimination of terms for
  dimensionality reduction, in: 2009 Ninth International Conference on
  Intelligent Systems Design and Applications, 2009, pp. 547--552.
\newblock \href {https://doi.org/10.1109/ISDA.2009.9}
  {\path{doi:10.1109/ISDA.2009.9}}.

\bibitem{Olive2017}
D.~J. Olive, \href{https://doi.org/10.1007/978-3-319-68253-2_6}{Principal
  Component Analysis}, Springer International Publishing, Cham, 2017, pp.
  189--217.
\newblock \href {https://doi.org/10.1007/978-3-319-68253-2_6}
  {\path{doi:10.1007/978-3-319-68253-2_6}}.
\newline\urlprefix\url{https://doi.org/10.1007/978-3-319-68253-2_6}

\bibitem{Bingham:2001:RPD:502512.502546}
E.~Bingham, H.~Mannila, \href{http://doi.acm.org/10.1145/502512.502546}{Random
  projection in dimensionality reduction: Applications to image and text data},
  in: Proceedings of the Seventh ACM SIGKDD International Conference on
  Knowledge Discovery and Data Mining, KDD '01, ACM, New York, NY, USA, 2001,
  pp. 245--250.
\newblock \href {https://doi.org/10.1145/502512.502546}
  {\path{doi:10.1145/502512.502546}}.
\newline\urlprefix\url{http://doi.acm.org/10.1145/502512.502546}

\bibitem{1053964}
T.~Cover, P.~Hart, Nearest neighbor pattern classification, IEEE Transactions
  on Information Theory 13~(1) (1967) 21--27.
\newblock \href {https://doi.org/10.1109/TIT.1967.1053964}
  {\path{doi:10.1109/TIT.1967.1053964}}.

\bibitem{708428}
M.~A. Hearst, S.~T. Dumais, E.~Osuna, J.~Platt, B.~Scholkopf, Support vector
  machines, IEEE Intelligent Systems and their Applications 13~(4) (1998)
  18--28.
\newblock \href {https://doi.org/10.1109/5254.708428}
  {\path{doi:10.1109/5254.708428}}.

\bibitem{breiman2017classification}
L.~Breiman, Classification and regression trees, Routledge, 2017.

\bibitem{casilari2016analysis}
E.~Casilari, J.~A. Santoyo-Ram{\'o}n, J.~M. Cano-Garc{\'\i}a, Analysis of a
  smartphone-based architecture with multiple mobility sensors for fall
  detection, PLoS one 11~(12) (2016) e0168069.

\bibitem{sztyler2016body}
T.~Sztyler, H.~Stuckenschmidt, On-body localization of wearable devices: An
  investigation of position-aware activity recognition, in: Pervasive Computing
  and Communications (PerCom), 2016 IEEE International Conference on, IEEE,
  2016, pp. 1--9.

\bibitem{Mafrur2015}
R.~Mafrur, I.~G.~D. Nugraha, D.~Choi,
  \href{https://doi.org/10.1186/s13673-015-0049-7}{Modeling and discovering
  human behavior from smartphone sensing life-log data for identification
  purpose}, Human-centric Computing and Information Sciences 5~(1) (2015) 31.
\newblock \href {https://doi.org/10.1186/s13673-015-0049-7}
  {\path{doi:10.1186/s13673-015-0049-7}}.
\newline\urlprefix\url{https://doi.org/10.1186/s13673-015-0049-7}

\bibitem{Krawczyk2016}
B.~Krawczyk, \href{https://doi.org/10.1007/s13748-016-0094-0}{Learning from
  imbalanced data: open challenges and future directions}, Progress in
  Artificial Intelligence 5~(4) (2016) 221--232.
\newblock \href {https://doi.org/10.1007/s13748-016-0094-0}
  {\path{doi:10.1007/s13748-016-0094-0}}.
\newline\urlprefix\url{https://doi.org/10.1007/s13748-016-0094-0}

\bibitem{chawla2002smote}
N.~V. Chawla, K.~W. Bowyer, L.~O. Hall, W.~P. Kegelmeyer, Smote: synthetic
  minority over-sampling technique, Journal of artificial intelligence research
  16 (2002) 321--357.

\bibitem{ARNABOLDI20173}
V.~Arnaboldi, M.~G. Campana, F.~Delmastro, E.~Pagani,
  \href{http://www.sciencedirect.com/science/article/pii/S1574119216301365}{A
  personalized recommender system for pervasive social networks}, Pervasive and
  Mobile Computing 36 (2017) 3 -- 24, special Issue on Pervasive Social
  Computing.
\newblock \href {https://doi.org/https://doi.org/10.1016/j.pmcj.2016.08.010}
  {\path{doi:https://doi.org/10.1016/j.pmcj.2016.08.010}}.
\newline\urlprefix\url{http://www.sciencedirect.com/science/article/pii/S1574119216301365}

\bibitem{7509388}
D.~Chatzopoulos, M.~Ahmadi, S.~Kosta, P.~Hui, Openrp: a reputation middleware
  for opportunistic crowd computing, IEEE Communications Magazine 54~(7) (2016)
  115--121.
\newblock \href {https://doi.org/10.1109/MCOM.2016.7509388}
  {\path{doi:10.1109/MCOM.2016.7509388}}.

\bibitem{8024034}
D.~Chatzopoulos, M.~Ahmadi, S.~Kosta, P.~Hui, Flopcoin: A cryptocurrency for
  computation offloading, IEEE Transactions on Mobile Computing 17~(5) (2018)
  1062--1075.
\newblock \href {https://doi.org/10.1109/TMC.2017.2748133}
  {\path{doi:10.1109/TMC.2017.2748133}}.

\bibitem{Chatzopoulos:2016:RRR:2964284.2967233}
D.~Chatzopoulos, P.~Hui,
  \href{http://doi.acm.org/10.1145/2964284.2967233}{Readme: A real-time
  recommendation system for mobile augmented reality ecosystems}, in:
  Proceedings of the 2016 ACM on Multimedia Conference, MM '16, ACM, New York,
  NY, USA, 2016, pp. 312--316.
\newblock \href {https://doi.org/10.1145/2964284.2967233}
  {\path{doi:10.1145/2964284.2967233}}.
\newline\urlprefix\url{http://doi.acm.org/10.1145/2964284.2967233}

\end{thebibliography}

\end{document}